\documentclass{acm_proc_article-sp}

\usepackage{amssymb}
\usepackage{graphicx}
\usepackage{verbatim}

\usepackage{epstopdf}

\usepackage{tikz}
\usetikzlibrary{positioning}
\usepackage{natbib}
\bibpunct{[}{]}{;}{a}{,}{,}

\usepackage{soul}

\begin{document}

\title{Learning measures of semi-additive behaviour
}

\numberofauthors{8} 

\author{
\alignauthor
Hamidreza Chinaei\titlenote{Work done while at IBM Canada.}\\
       \affaddr{Department of Computer Science}\\
       \affaddr{University of Toronto}\\
       \affaddr{Toronto, ON, Canada}\\
       \email{chinaei@cs.toronto.edu}
% 2nd. author
\alignauthor
Mohsen Rais-Ghasem\\
\affaddr{Business Analytics Team}\\
       \affaddr{IBM Canada}\\
%       \affaddr{}\\
       \affaddr{Ottawa, ON, Canada}\\
       \email{mohsen.rais-ghasem@ca.ibm.com }
% 2nd. author
\alignauthor
Frank Rudzicz\\
\affaddr{Department of Computer Science}\\
       \affaddr{University of Toronto}\\
%       \affaddr{}\\
       \affaddr{Toronto, ON, Canada}\\
       \email{frank@cs.toronto.edu }
}

\maketitle

\begin{abstract}
\begin{quote}
In business analytics, \emph{measure} values, such as sales numbers or volumes of cargo transported, are often summed along values of one or more corresponding \emph{categories}, such as time or shipping container. However, not every measure should be added by default (e.g., one might more typically want a {\em mean} over the heights of a set of people); similarly, some measures should only be summed within certain constraints (e.g., population measures need not be summed over years). In systems such as Watson Analytics, the exact additive behaviour of a measure is often determined by a human expert. In this work, we propose a small set of features for this issue. We use these features in a case-based reasoning approach, where the system suggests an aggregation behaviour, with 86\% accuracy in our collected dataset. 
\end{quote}
\end{abstract}

\section{Introduction}
\begin{figure}[!t]
\centering
\centering
\includegraphics[scale=0.60]{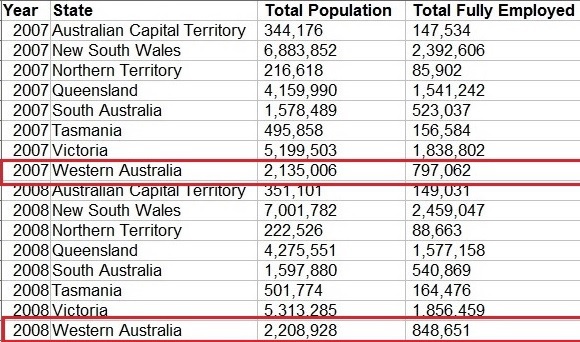}
\caption{\label{fig:australia-states} The snapshot of a spreadsheet containing statistics for employment in Australian states.}
\label{fig:employment}
\end{figure}

\begin{figure}[!t]
\centering
\includegraphics[scale=0.87]{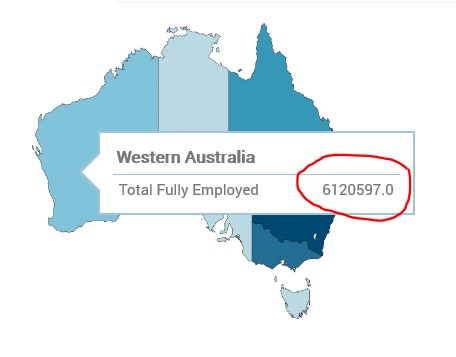}
\caption{\label{fig:australia-states-results} The result of a default aggregation on the spreadsheet of statistics for employment in Australian states for the user question: \textit{"What are the values of Total Fully Employed by State?}.}
\label{fig:employment-results}
\end{figure}

\begin{figure}[!h]
\begin{center}
\includegraphics[scale=0.65]{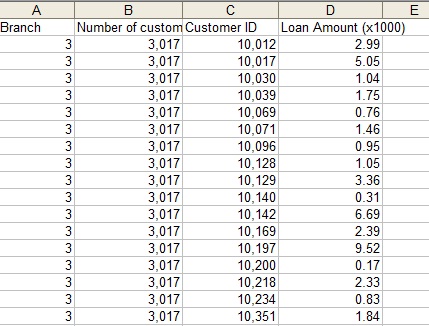}
\caption{\label{fig:bank-loan-data} A snapshot of the bank loan spreadsheet.}
\end{center}
\end{figure}

\begin{figure}[!h]
\begin{center}
\includegraphics[scale=0.55]{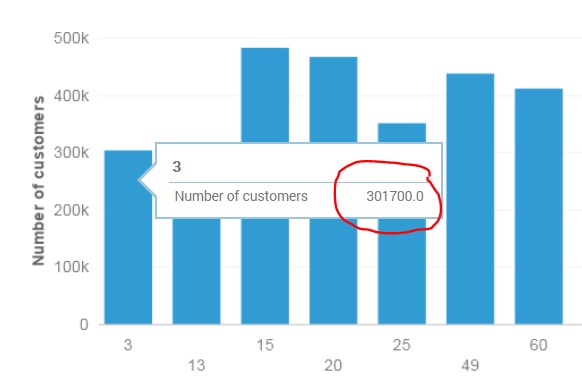}
\caption{\label{fig:bank-loan-aggregate} The result of a default aggregation on the bank loan spreadsheet, for the user's question: \textit{"Number of customers for each branch?}.}
\end{center}
\end{figure}

In business analytics, \emph{measure} values are often summed, but often other aggregation measures, such as means or variances, is more appropriate. For example, the dataset in Figure \ref{fig:employment} contains employment statistics for Australian states for 2007 and 2008; clearly, while summing over partitions by year may be appropriate, but summing over the entire column is not. The result of a default aggregation (sum) over these data is shown in Figure~\ref{fig:employment-results} given the user question: "What are the values of Total Fully Employed by State?".
Figure~\ref{fig:bank-loan-data} shows another example dataset that tracks information about loan requests received by branches of a bank. The number of clients is incorrectly added in a branch by branch analysis (see Figure~\ref{fig:bank-loan-aggregate}). Here, adding the loan amounts for each branch makes sense, but adding the number of clients does not.
This problem is usually avoided by defining the correct default aggregation behaviour ahead of time, through data modeling techniques, which often requires hand-tuning each variable in a dataset. 

In this work, we propose an alternative approach. We introduce a small set of features and develop them in a CBR system to suggest better default aggregations and hence eliminate or at least accelerate prior, data-specific modeling. This is done by extracting potentially useful features from the data and by applying case-based reasoning (CBR)~\citep{SchankInsideCBR1989,Kolodner1993case}. 

Our CBR system is fed a few use cases representing an aggregation context (so called \emph{known cases}), consisting of (i) a \emph{measure} item whose default aggregation is being learned or queried (e.g. ``Total Fully Employed'' in Figure~\ref{fig:employment}); (ii) one or more \emph{category} items for which the given measure has repeated values that need to be aggregated (e.g. ``State'' and ``Year'' in Figure~\ref{fig:employment}); (iii) the expected default aggregate which is one of \textit{sum, average}, or \textit{last-period}. 

In this work, we propose a small set of features that, when used in our CBR system, improve the learning of appropriate aggregate actions up to 86\% accuracy. In the rest of this paper, we first briefly explain CBR and our CBR system architecture in Section~\ref{sec:cbr}. We describe our extracted features for our CBR package in Section~\ref{sec:features}. The similarity measure for each feature is described in Section~\ref{sec:sim}. We then go through our empirical settings in Section~\ref{sec:experiments}. Finally, we conclude and address future directions in Section~\ref{sec:conclusion}.

\section{CBR architecture}\label{sec:cbr}
CBR has been applied successfully in many practical domains ~\citep{Lamontagne2014ICCBR,Jalali14,ChenFWP14,Dong14,Freyne2010,Jurisica1997}. 
The principal idea in CBR is to draw parallels between a new case and those that have been already~solved. 

\begin{figure}[!h]
\begin{center}
\includegraphics[scale=0.35]{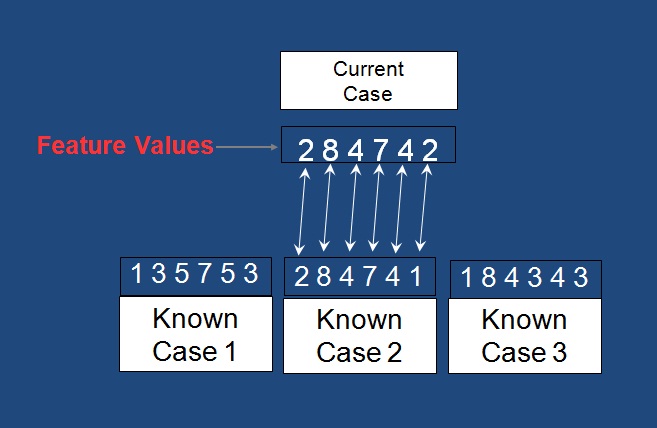}
\caption{\label{fig:cbr} Case-based reasoning (CBR): In CBR a new case (current case) is compared against existing cases using a set of extracted features.} 
\end{center}
\end{figure}

Figure~\ref{fig:cbr} shows an example case represented by feature values that need to be extracted. The action for each known case is defined and learned. The best action for the current case is selected using the most similar cases to the current one. That is, the current case is compared to the known cases using feature similarity measures and the case features, then the action for the most similar case, among all known cases, is selected for and applied to the current case. 

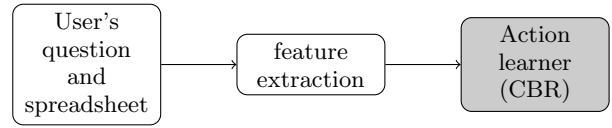
\begin{figure}[!h]
\begin{center}
\begin{tikzpicture}[box/.style={draw,rounded corners,text width=1.75cm,align=center}]
\node[box] (a0) {User's question and spreadsheet};
\node[box,right=of a0] (a) {feature\\extraction};
\node[box,right=of a, fill=black!20] (b) {Action learner\\(CBR)};

\draw[->] (a0) -- (a);
\draw[->] (a) -- (b);
\end{tikzpicture}
\end{center}
\caption{\label{fig:architecture} The architecture of our CBR system.}
\end{figure}
Figure~\ref{fig:architecture} demonstrates the architecture of our developed aggregate learner. The user's spreadsheet is submitted to the system followed by the user's question(s). The system extracts the features for the spreadsheet particularly for \textit{category} and \textit{measure} columns. These features are described in Section~\ref{sec:features}. The action learner component receives the features and finds the aggregate action for each measure column based on CBR.

\section{Extracted Features}\label{sec:features}
At the heart of our system, and any CBR system, lies a carefully selected \emph{feature set} that is used to measure the similarity of the current case to known cases, as depicted in Figure~\ref{fig:cbr}. The role that each feature plays in the overall similarity measurement (i.e., the feature weights) can be assigned manually or automatically using a number of optimization techniques similar to those of ~\cite{Lamontagne2014}.

Table~\ref{tab:case-representation} demonstrates the representation of a case using the extracted features for the bank loan example in Figure~\ref{fig:bank-loan-data}.
\begin{table}[!h]
\caption{Example of a case representation in our CBR system.}\label{tab:case-representation}
\centering
\scriptsize
\begin{tabular}{|ll|}
\hline
dataset: & bankloan.xls\\
measure column: & ``Loan Amount (x1000)''\\
category column: & ``Branch''\\
\hline
concepts (measure column): & \textit{[metric, monetary]}\\
concepts (category column): & \textit{[attribute]}\\
association type (category to measure): & \textit{one-to-many}\\
averaged CoV: & \textit{1.37}\\
\hline
measure aggregate action: &sum\\
\hline
\end{tabular}
\end{table}
Our feature set consists of factors that we found most useful in establishing the default aggregation behaviour, most notably the following features:
\begin{enumerate}
\item \emph{Semantic annotations} of measure columns and category columns, similar to column \emph{concepts} as defined in~\cite{Rais-Ghasem2013};
\item \emph{Association type} between category columns and measure columns;
\item \emph{Coefficient of variation (\textit{CoV})} as an indication of trends of measure values in the context of given~categories.
\end{enumerate}

\subsection*{Semantic annotations}
For the semantic annotations of measures and categories (feature 1), consider the measure column ``Loan Amount (x1000)'', and the category column ``Branch''. The semantic annotations of the measure column are captured as \textit{metric} and \textit{monetary}, and the semantics annotations of the category column is captured as \textit{attribute}~\citep{Rais-Ghasem2013}. In particular, we used a designation shared by many metrics in our known cases in our CBR system, and specifically avoided depending on explicit semantic knowledge about items.

\subsection*{Association types}\label{sec:correlation}
A simple analysis of data values in Figure~\ref{fig:bank-loan-data} reveals that values of column ``A'' (``Branch'') and ``B'' (``Customer ID'') have a \emph{one-to-one association type}, whereas the values of column ``A'' and ``D'' (Loan Amount (x1000)) have a \emph{one-to-many association type}. That is, for each ``Branch'' number there is only one value for ``number of customer'', thus a \textit{one-to-one} association type. Note that the association type feature between a category column and measure column can be assigned four different values: \textit{one-to-one}, \textit{one-to-many}, \textit{many-to-one}, and \textit{many-to-many}.

As useful as this might be (i.e., it may suggest a non-additive behaviour for \textit{B} in the context of \textit{A}), it is not sufficient to identify the correct behaviour for the employment use case in Figure~\ref{fig:employment}. That is, for each category column ``Year'' (or ``State'') there is many ``Total Fully Employed'' values, thus \textit{many-to-many} association type. However, addition is not the correct aggregate behaviour for ``'Total Fully Employed' over ``Year''. We thus introduce the third feature that can potentially solve this limitation.

\subsection*{Value trends}
We observe that trends of value dynamics are often useful, especially in combination with the aforementioned association type and semantic annotation of category (e.g., ``Year'' is a temporal category, whereas ``Branch'' is not).

\begin{figure*}[!t]
\begin{center}
\includegraphics[scale=0.65]{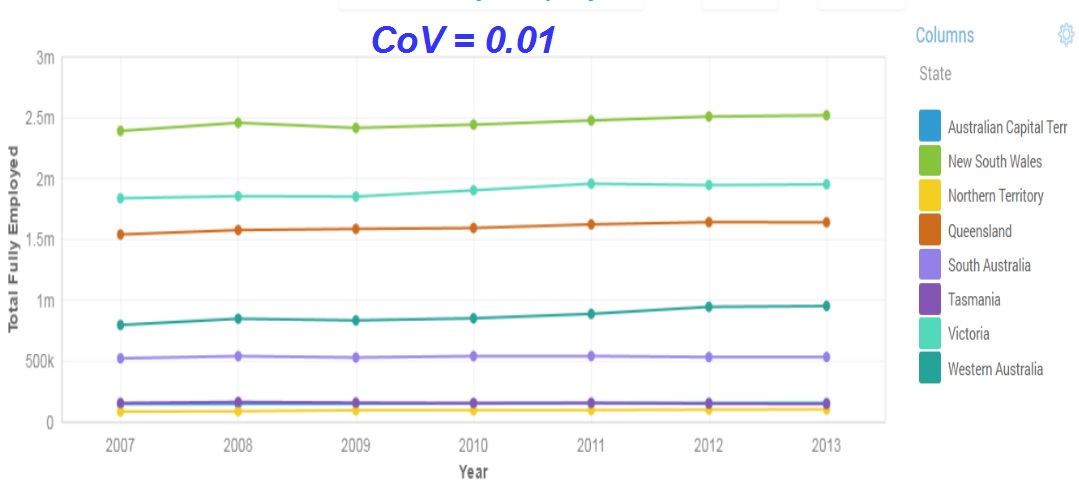}
\caption{\label{fig:australia-states-trend} The trend of Australia states data. The \textit{CoV} for ``Totally Fully Employed'' over ``Year'' is close to~0. }
\end{center}
\end{figure*}

\begin{figure}[!h]
\begin{center}
\includegraphics[scale=0.75]{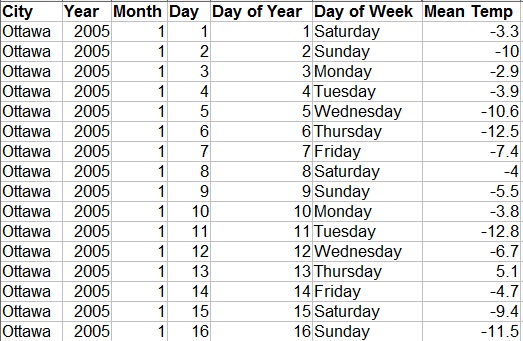}
\caption{\label{fig:weather-data} The temperature readings spreadsheet.}
\end{center}
\end{figure}

\begin{figure*}[!t]
\begin{center}
\includegraphics[scale=0.65]{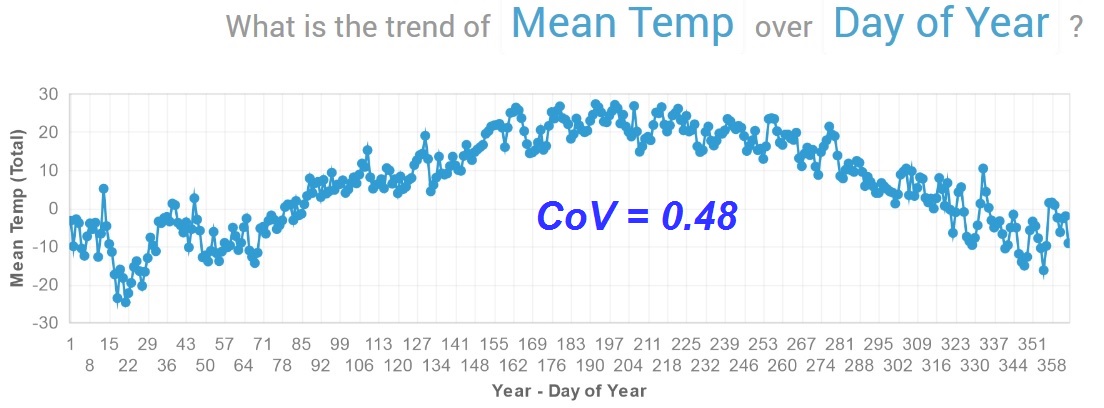}
\caption{\label{fig:weather-trend} The trend of temperature data. The \textit{CoV} for ``Mean Temp'' over ``Days of Year'' is between 0 and~1.}
\end{center}
\end{figure*}

\begin{figure*}[!t]
\begin{center}
\includegraphics[scale=0.65]{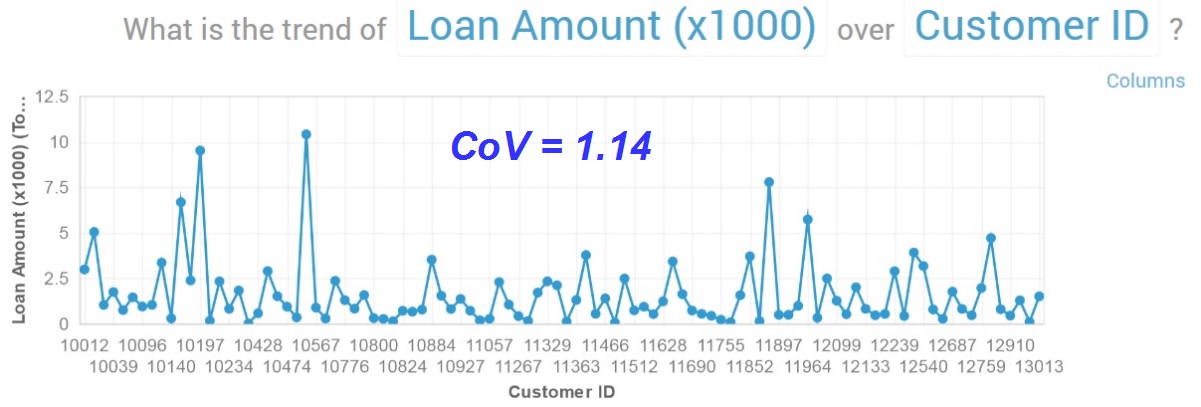}
\caption{\label{fig:bankloan-trend} The trend of bank loan data. The \textit{CoV} for ``Loan Amount'' over ``Customers'' is close to 1.}
\end{center}
\end{figure*}

To quantify data trends, we use the \emph{coefficient of variation (\textit{CoV})}:
\[\mbox{\textit{CoV}} = \sigma/\mu\]
where $\sigma$ is the standard deviation and $\mu$ is the mean. Here, \textit{CoV} acts as normalized standard variation. The lower the \textit{CoV} measure for a column of data, the greater the trend tended to be in early empirical tests. This can be seen in Figure~\ref{fig:australia-states-trend} to Figure~\ref{fig:bankloan-trend}. % which is explained below. 

Figure~\ref{fig:australia-states-trend} shows the \textit{CoV} for the employment status. The figure shows that the \textit{CoV} for ``Totally Fully Employed'' over ``Year'' is close to 0. Note that the average of \textit{CoV} over different states is 0.01.

We have done a similar analysis for data in Watson Analytics that contains temperatures for Canadian cities. Figure~\ref{fig:weather-data} shows a snapshot of this spreadsheet. Figure~\ref{fig:weather-trend} shows the \textit{CoV} for the temperature, averaged over the category column (cities). This figure shows that the \textit{CoV} for ``Mean Temperature'' over ``Day of Year'' is between 0 and 1. That is, the average \textit{CoV} over different cities is 0.48. In this analysis, to get sensible \textit{CoV} values we had to transform the measure values to positive (a negative measure value occurs in the column). To do so, we add the absolute value of the minimum value of the measure column to all values in that measure column. %\FR{WHY?}

Finally, we follow a similar analysis over bank loan data, whose \textit{CoV} is shown in Figure~\ref{fig:bankloan-trend}. Here, \textit{CoV} for ``Loan Amount'' over ``Customer ID'' is close to 1 and the average \textit{CoV} over different branches is 1.14.

Thus an interesting pattern emerges, as summarized in Table~\ref{tab:cov-pattern}. \textit{CoV} for measure values repeated for a category (e.g., various employment figures for different years) had strong associations with expected aggregation functions. Using \textit{CoV} in combination with other features to propose default aggregation automatically, in general, within IBM Watson Analytics. 

\begin{table}[!h]
\caption{The \textit{CoV} patterns for aggregate learning.}\label{tab:cov-pattern}
\centering
\begin{tabular}{|l|l|}
\hline
Measure \textit{CoV} values & Default Aggregation\\
\hline
Values close to 0 & \textit{last-period}\\
Values close to 0.5 & \textit{average}\\
Values close to 1 & \textit{sum}\\
\hline
\end{tabular}

\end{table}

\section{Feature similarity function}\label{sec:sim}
The similarity function for each feature is defined as follows.

\noindent \textbf{Feature 1 (column concepts):} \[Sim(f_1(c_1),f_1(c_2)) = \frac{card(~f_1(c_1) \cap f_1(c_2)~) * 2}{card(~f_1(c_1)~)+ card(~f_1(c_2)~)}\]
where $c_1$ and $c_2$ are case 1 and case 2, respectively; and $f_1(c_i)$ returns the value of feature one for case $i$. Note that feature 1 is the list of column concepts. In the above formula, $\mbox{\textit{card()}}$ returns the cardinality of each set. Effectively, the similarity is the number of common concepts (feature 1 values) for case 1 and case 2 divided by the total number of column concepts. Clearly, if the two cases share no concept, the similarity is 0, and if the two cases are identical, the similarity is 1.

\noindent \textbf{Feature 2 (association type between columns):} This similarity measure is defined by our domain experts as follows:
If the association types are the same for the two cases then similarity is 1. 
The similarity between \textit{many-to-many} and \textit{many-to-one} association type is 0.5. 
The similarity between \textit{one-to-one} and anything else (\textit{one-to-many} or \textit{many-to-many}) is 0. \\

\noindent \textbf{Feature 3 (value trend):} Here, the sigmoid function defines similarity between 0 and 1.
\[Sim(f_3 (c_1), f_3(c_2))=\frac{1}{1+e^{-x}}\]
where $f_3(c_i)$ is the \textit{CoV} values (feature 3) for case $c_i$; $x$ computes how close $CoV$ values for case 1 and case 2 are, defined as : $1/(|f_3(c_1)-f_3(c_2)|)$ . Recall that \textit{CoV} is the coefficient of variation as defined in Section~\ref{sec:features}.

The similarity between two cases is calculated based on the similarity of their feature values (the three introduced feature-based similarities above). Each feature-based similarity produces values between 0 and 1, and the total similarity function is the weighted sum of the three feature-based similarity (in which the weights are set to 1 in this work). Thus the total similarity is a value between 0 and 1.

\section{Experiments}\label{sec:experiments}
We collected about 100 such use cases from IBM Watson Analytics and used 65\% of them to be used as the known cases in our CBR system and the remaining 35\% to evaluate our approach. Each case includes a question posed on a 2-dimensional spreadsheet of data in which we explicitly denote the \textit{category} and \textit{measure} columns, done manually by our domain experts. We then extracted the three features described in Section~\ref{sec:features} for all the known cases in our CBR system. We then tested our approach using the remaining evaluation data. 

Our evaluations use a `majority voting' scheme in which the top $k$ most similar cases to the test case are selected and the aggregate action for the test case is selected based on the most frequent aggregate action of these $k$ most similar cases. In the extreme, if $k=1$, the action of the most similar `known' case is picked as the aggregate action for the test case. For the majority vote, we empirically set $k=3$, given that we find no significant difference in sweeping $k$ from 3 to 6. Further increases of $k>6$ results in in lower~accuracy.

Table~\ref{tab:results} shows that the measure column's \textit{CoV} leads to the highest accuracy when each feature is used in isolation. Furthermore, when all the three features are used together, the approach achieves 86\% accuracy using majority vote.

\begin{table}[!h]
\caption{The accuracy captured for the testing data using CBR and three different features.}\label{tab:results}
\centering
\begin{tabular}{|l|c|}
\hline
Method & Accuracy \\
\hline
F1 (columns' concept)& 30\%\\
F2 (columns' association type)& 30\%\\
F3 (measure column's \textit{CoV})& 63\%\\
All the three features & 83\%\\
All the three features with majority vote & \textbf{86\%}\\
\hline
\end{tabular}

\end{table}

\begin{figure}[!t]
\centering
\includegraphics[scale=0.60]{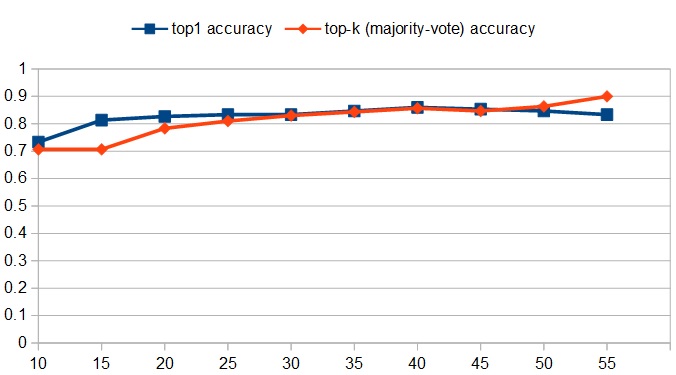}
\caption{Trend of accuracy by increasing the number of training cases.}
\label{fig:trend-results}
\end{figure}

Furthermore, we calculate the accuracy trend over known cases. Figure~\ref{fig:trend-results} shows by increasing the known cases, while the testing data remains the same, our approach achieves better accuracy. Specifically, the accuracy increases from 70\% to 90\% by increasing the number of known cases from 10 to 55. 

Notice that, in this work, we use the same feature weights for all the features (all feature weight are 1). In the future, we are going to learn feature weights using optimization techniques similar to those used by~\cite{Lamontagne2014}.

Finally, we calculate the system-level elapsed time in milliseconds for different tasks, shown in Table~\ref{tab:elapsed-time}. The feature selection task (for all known cases) takes less than two minutes by a ThinkPad (T410s) with i5-2.4GHz and 8GB of RAM, and using Win 7 Professional 64bit. 
\begin{table}[!h]
\caption{The system elapsed time in milliseconds for different kind of tasks.}\label{tab:elapsed-time}
\small
\centering
\begin{tabular}{|l|l|}
\hline
Measure \textit{CoV} values & Default Aggregation\\
\hline
Feature Extraction (known cases) & 83594\\
Aggregate Learning (average) & 8.46\\
Aggregate Learning (range) & 1 $< t <$ 28\\
\hline
\end{tabular}
\end{table}

The average time for learning aggregate actions of each test case is about 8 milliseconds. The learning time varies for each test case (between 1 and 28 milliseconds) since the feature extraction for each test case is done at run time and, in particular, calculating the \textit{CoV} highly depends on the number of columns in the data.

\subsection{Examples of experimental results}
Here we isolate three test cases and their most similar cases as identified by our approach, shown in Table~\ref{tab:sample-results}. In the first two cases, our approach has been able to find the correct aggregate action, however our approach fails in the third case to learn the correct aggregate action. 

Our approach does not learn the correct default aggregate action for test case 3 (code coverage of each component). The suggested action by our approach is \textit{sum} instead of \textit{last-period}. The code coverage spreadsheet keeps the percentage of code that has been tested over time for each component of a software. Thus, the correct aggregate action is taking \textit{last-value}. We believe that this error is caused small number of rows in the code coverage spreadsheet. In particular, the calculated \textit{CoV} for data in the code coverage dataset does not represent the trend of coverage values well enough. 
\begin{table*}
\caption{Examples of our experimental data and results.}\label{tab:sample-results}
\centering
%\scriptsize
\begin{tabular}{|l|c|}
\hline
\multicolumn{2}{ |c| }{Test case 1} \\
\hline
Data set & American Time Use Survey.csv\\
Question & What are values of sleeping time over gender?\\
Category column & [Gender]\\
Measure column & [Sleeping]\\
\hline
\multicolumn{2}{ |c| }{Most similar known case}\\
\hline
Data set & GL Budget.xls\\
Question & What are values of budget over account?\\
category column & [Account-Number]\\
measure column & [Period-Budget-Amount]\\
aggregate action & \textcolor{blue}{[average]} \\
\hline
\hline
\multicolumn{2}{ |c| }{Test case 2} \\
\hline
Data set & [Ticket-Sales-by-Section-with-Geo.csv]\\
Question & What are total tickets purchased for each event ?\\
Category column & [EventName] \\
Measure column & [Total Tickets Purchased]\\
\hline
\multicolumn{2}{ |c| }{Most similar known case}\\
\hline
Data set & [Ticket-Sales-by-Section-with-Geo.csv]\\
Question & What are total price of purchased tickets for each event ?\\
category column & [EventName] \\
measure column &[Total Ticket Purchase Price]\\
aggregate action & \textcolor{blue}{[sum]}\\
\hline
\hline
\multicolumn{2}{ |c| }{Test case 3} \\
\hline
Data set & Sonar Results over Time.xls\\
Question & What is code coverage for each component?\\
Category column & [Component]\\
Measure column & [Coverage]\\
\hline
\multicolumn{2}{ |c| }{Most similar known case}\\
\hline
Data set & World Mortality.csv\\
Question & What are the number of deaths by cause?\\
category column & [Cause]\\
measure column & [Number of Deaths]\\
aggregate action & \textcolor{red}{[sum]} \\
\hline
\end{tabular}
\end{table*}

\section{Conclusion}\label{sec:conclusion}
In this work, we proposed a method for learning semi-additive behaviour in data analytics tools, chiefly IBM Watson Analytics in which these experiments were performed. Specifically, we introduced three features that are used in case-based reasoning for learning aggregate actions in business analytics. In particular, the data trend of a variable is an important factor that can be used for learning the default aggregate of that variable. We calculated the data trend using \emph{coefficient of variation (\textit{CoV})}, whose use can empirically increase the accuracy up to 30\%. Overall, our approach is able to learn the default aggregate of our testing data with up to 90\% accuracy. In this work, we used the same feature weights for all the features (all feature weights are 1.0). In the future, we intend to learn feature weights using optimization techniques such as linear programming, and we are collecting more test cases to experiment our approach on larger data sets. 

\section*{Acknowledgment}
This research is supported in part by a contribution from IBM Canada.

\balancecolumns

\bibliographystyle{apalike}

\bibliography{Learning-Measures-of-Semi-Additive-Behaviour}

\end{document}